\documentclass[
twocolumn,
]{ceurart}

\usepackage{tikz}
\usetikzlibrary{positioning}
\usepackage{fontawesome}
\usepackage{array}
\newcolumntype{P}[1]{>{\centering\arraybackslash}p{#1}}
\newcolumntype{M}[1]{>{\centering\arraybackslash}m{#1}}
\usepackage[many]{tcolorbox}

\usepackage{xurl}

\begin{document}\sloppy

\copyrightyear{2021}
\copyrightclause{Copyright for this paper by its authors.
  Use permitted under Creative Commons License Attribution 4.0
  International (CC BY 4.0).}

\conference{Joint Proceedings of the ACM IUI 2021 Workshops,
  April 13--17, 2021, College Station, USA}

\title{A Study on Fairness and Trust Perceptions in Automated Decision Making}

\author[1]{Jakob Schoeffer}[%
orcid=0000-0003-3705-7126,
]
\ead{jakob.schoeffer@kit.edu}
\address[1]{Karlsruhe Institute of Technology (KIT), Germany}

\author[1]{Yvette Machowski}
[orcid=0000-0002-9271-6342,]
\ead{yvette.machowski@alumni.kit.edu}

\author[1]{Niklas Kuehl}[%
orcid=0000-0001-6750-0876,
]
\ead{niklas.kuehl@kit.edu}

\begin{abstract}
Automated decision systems are increasingly used for consequential decision making---for a variety of reasons.
These systems often rely on sophisticated yet opaque models, which do not (or hardly) allow for understanding \emph{how} or \emph{why} a given decision was arrived at.
This is not only problematic from a legal perspective, but non-transparent systems are also prone to yield undesirable (e.g., unfair) outcomes because their sanity is difficult to assess and calibrate in the first place.
%
%
In this work, we conduct a study to evaluate different attempts of explaining such systems with respect to their effect on people's perceptions of fairness and trustworthiness towards the underlying mechanisms.
%
A pilot study revealed surprising qualitative insights as well as preliminary significant effects, which will have to be verified, extended and thoroughly discussed in the larger main study.
\end{abstract}

\begin{keywords}
  Automated Decision Making \sep
  Fairness \sep
  Trust \sep
  Transparency \sep
  Explanation \sep
  Machine Learning
\end{keywords}

\maketitle

\section{Introduction}
Automated decision making has become ubiquitous in many domains such as hiring \citep{kuncel2014hiring}, bank lending \citep{townson2020ai}, grading \citep{satariano2020british}, and policing \citep{heaven2020predictive}, among others.
As automated decision systems (ADS) are used to inform increasingly high-stakes consequential decisions, understanding their inner workings is of utmost importance---and undesirable behavior becomes a problem of societal relevance.
The underlying motives of adopting ADS are manifold: They range from cost-cutting to improving performance and enabling more robust and objective decisions \citep{kuncel2014hiring,harris2005automated}. 
One widespread assumption is that ADS can also avoid human biases in the decision making process \citep{kuncel2014hiring}.
However, ADS are typically based on artificial intelligence (AI) techniques, which, in turn, generally rely on historical data.
If, for instance, this underlying data is biased (e.g., because certain socio-demographic groups were favored in a disproportional way in the past), an ADS may pick up and perpetuate existing patterns of unfairness \citep{barocas2018fairness}.
Two prominent examples of such behavior from the recent past are the discrimination of black people in the realm of facial recognition \citep{buolamwini2018gender} and recidivism prediction \citep{angwin2016machine}.
These and other cases have put ADS under enhanced scrutiny, jeopardizing trust in these systems.

In recent years, a significant body of research has been devoted to detecting and mitigating unfairness in automated decision making
\citep{barocas2018fairness}.
Yet, most of this work has focused on formalizing the concept of fairness and enforcing certain statistical equity constraints, often without explicitly taking into account the perspective of individuals affected by such automated decisions.
In addition to how researchers may define and enforce fairness in technical terms,
we argue that it is vital to understand people's \emph{perceptions} of fairness---vital not only from an ethical standpoint but also with respect to facilitating trust in and adoption of (appropriately deployed) socio-technical systems like ADS.
\citet{srivastava2019mathematical}, too, emphasize the need for research to gain a deeper understanding of people's attitudes towards fairness in ADS.

A separate, yet very related, issue revolves around how to \emph{explain} automated decisions and the underlying processes to affected individuals so as to enable them to appropriately assess the quality and origins of such decisions.
\citet{srivastava2019mathematical} also point out that subjects should be presented with more information about the workings of an algorithm and that research should evaluate how this additional information influences people's fairness perceptions.
%
In fact,
the EU General Data Protection Regulation (GDPR)\footnote{\url{https://eur-lex.europa.eu/eli/reg/2016/679/oj} (last accessed Jan 3, 2021)}, for instance, requires to disclose ``the existence of automated decision-making, including [\dots] meaningful information about the logic involved [\dots]'' to the ``data subject''. 
Beyond that, however, such regulations remain often vague and little actionable.
%
%
To that end, we conduct a study to examine in more depth the effect of different explanations on people's perceptions of fairness and trustworthiness towards the underlying ADS in the context of lending, with a focus on
\begin{itemize}
    \item the amount of information provided,
    \item the background and experience of people,
    \item the nature of the decision maker (human vs. automated).
\end{itemize}


\section{Background and Related Work}
It is widely understood that AI-based technology can have undesirable effects on humans.
%
As a result, topics of fairness, accountability and transparency have become important areas of research in the fields of AI and human-computer interaction (HCI), among others.
In this section, we provide an overview of relevant literature and highlight our contributions.

\begin{table*}
\caption{Overview of related work.}
\label{tab:overview}
\resizebox{\textwidth}{!}{
    \begin{tabular}{M{1.5cm}  M{2.4cm}  M{2.4cm}  M{2.4cm}  M{2.4cm}  M{3cm}}
        \toprule
        \textbf{Reference} & \textbf{Explanation styles provided} & \textbf{Amount of provided information evaluated} & \textbf{Understandability tested} & \textbf{Computer / AI experience evaluated} &  \textbf{Human involvement in context considered} \\
        \midrule
        \citet{binns2018s} & distinct & no & single question & no &  no \\
        \midrule
        \citet{dodge2019explaining} & distinct & no & not mentioned & no & no \\
        \midrule
        \citet{lee2018understanding} & distinct & no & no & knowledge of algorithms & individual in management context \\
        \midrule
        \citet{lee2017algorithmic} & n/a due to study setup & no & no & programming / algorithm knowledge & group decision in fair division context \\
        \midrule
        \citet{wang2020factors} & distinct & partly & no & computer literacy & algorithmic decision, reviewed by group in crowdsourcing  context \\
        \midrule
        \midrule
        Our work & distinct and combined & yes & construct with multiple items & AI literacy & individual in provider-customer context \\
        \bottomrule
    \end{tabular}
}
\end{table*}

\paragraph{\textbf{Explainable AI}}
%
Despite being a popular topic of current research, explainable AI (XAI) is a natural consequence of designing ADS and, as such, has been around at least since the 1980s \citep{goebel2018explainable}.
Its importance, however, keeps rising as increasingly sophisticated (and opaque) AI techniques are used to inform evermore consequential decisions.
%
%
%
%
XAI is not only required by law (e.g., GDPR, ECOA\footnote{Equal Credit Opportunity Act: \url{https://www.consumer.ftc.gov/articles/0347-your-equal-credit-opportunity-rights} (last accessed Jan 3, 2021)}); \citet{eslami2019user}, for instance, have shown that users' attitudes towards algorithms change when transparency is increased.
When sufficient information is not presented, users sometimes rely too heavily on system suggestions \cite{bussone2015role}.
%
Yet, both quantity and quality of explanations matter: \citet{kulesza2013too} explore the effects of soundness and completeness of explanations on end users' mental models and suggest, among others, that oversimplification is problematic.
We refer to \citep{goebel2018explainable,molnar2020interpretable,adadi2018peeking} for more in-depth literature on the topic of XAI.

%
%
%

\paragraph{\textbf{Perceptions of fairness and trustworthiness}}
A relatively new line of research in AI and HCI has started focusing on \emph{perceptions} of fairness and trustworthiness in automated decision making.
For instance, \citet{binns2018s} and \citet{dodge2019explaining} compare fairness perceptions in ADS for four distinct explanation styles.
\citet{lee2018understanding} compares perceptions of fairness and trustworthiness depending on whether the decision maker is a person or an algorithm in the context of managerial decisions.
\citet{lee2017algorithmic} explore how algorithmic decisions are perceived in comparison to group-made decisions.
\citet{wang2020factors} combine a number of manipulations, such as favorable and unfavorable outcomes, to gain an overview of fairness perceptions.
An interesting finding by \citet{lee2019procedural} suggests that fairness perceptions decline for some people when gaining an understanding of an algorithm if their personal fairness concepts differ from those of the algorithm.
%
%
%
Regarding trustworthiness, \citet{kizilcec2016much}, for instance, concludes that it is important to provide the right amount of transparency for optimal trust effects, as both too much and too little transparency can have undesirable effects.

\paragraph{\textbf{Our contribution}}
We aim to complement existing work to better understand \emph{how much} of \emph{which} information of an ADS should be provided to \emph{whom} so that people are optimally enabled to understand the inner workings and appropriately assess the quality (e.g., fairness) and origins of such decisions.
Specifically, our goal is to add novel insights in the following ways: First, our approach combines multiple explanation styles in one condition, thereby disclosing varying amounts of information.
This differentiates our method from the concept of distinct individual explanations adopted by, for instance, \citet{binns2018s}.
%
%
We also evaluate the understandability of explanations through multiple items; and we add a novel analysis of the effect of people's AI literacy \cite{long2020ai} on their perceptions of fairness and trustworthiness.
%
%
%
%
%
Finally, we investigate whether perceptions of fairness and trustworthiness differ between having a human or an automated decision maker, controlling for the provided explanations.
For brevity, we have summarized relevant aspects where our work can complement existing literature in Table \ref{tab:overview}.
%
%

\section{Study Design and Methodology}\label{sec:study_design}

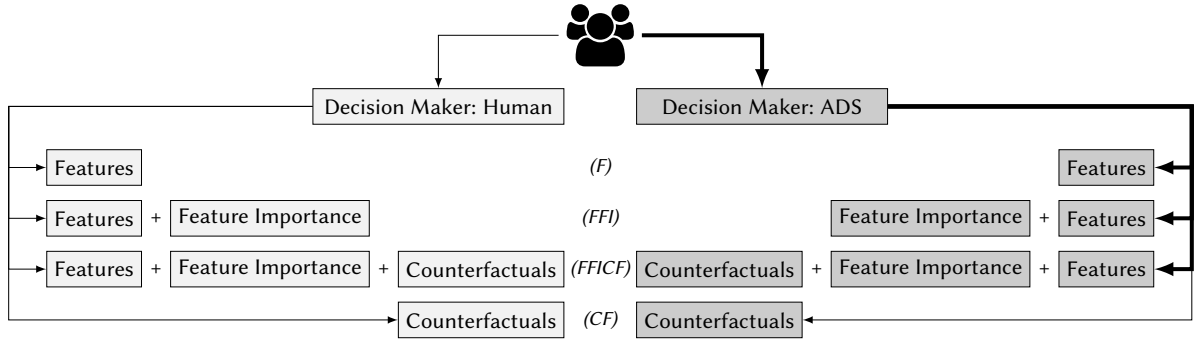
\begin{figure*}
\centering
\resizebox{\textwidth}{!}{
\begin{tikzpicture}  
  [scale=1] 
   
\node (n0) at (0,0) {\Huge \faUsers}; 

\node[style={rectangle, draw=black, fill=gray!10, minimum width=3.5cm, minimum height=0.5cm, text centered, anchor=east}] (n1) at (-0.5,-1) {Decision Maker: Human}; 

\node[style={rectangle, draw=black, fill=gray!40, minimum width=3.5cm, minimum height=0.5cm, text centered, anchor=west}] (n2) at (0.5,-1) {Decision Maker: ADS}; 


\node[style={rectangle, draw=black, fill=gray!10, minimum height=0.5cm, text centered, anchor=east}] (n3) at (-0.5,-4) {Counterfactuals};

\node[style={anchor=center, minimum height=0.5cm}] (txt1) at (0,-4) {\footnotesize{\textit{(CF)}}};

\node[style={rectangle, draw=black, fill=gray!10, minimum height=0.5cm, text centered, above=0.2cm of n3.north, anchor=south}] (n4) {Counterfactuals};

\node[style={left=0cm of n4.west}] (plus1) {+};

\node[style={rectangle, draw=black, fill=gray!10, minimum height=0.5cm, text centered, left=0cm of plus1}] (n5) {Feature Importance};

\node[style={left=0cm of n5.west}] (plus2) {+};

\node[style={rectangle, draw=black, fill=gray!10, minimum height=0.5cm, text centered, left=0cm of plus2}] (n6) {Features}; 

\node[style={above=0.2cm of txt1.north, anchor=south, minimum height=0.5cm}] (txt2) {\footnotesize{\textit{(FFICF)}}};

\node[style={rectangle, draw=black, fill=gray!10, minimum height=0.5cm, text centered, above=0.2cm of n5.north, anchor=south}] (n7) {Feature Importance}; 

\node[style={left=0cm of n7.west}] (plus3) {+};

\node[style={rectangle, draw=black, fill=gray!10, minimum height=0.5cm, text centered, left=0cm of plus3}] (n8) {Features}; 

\node[style={above=0.2cm of txt2.north, anchor=south, minimum height=0.5cm}] (txt3) {\footnotesize{\textit{(FFI)}}};

\node[style={rectangle, draw=black, fill=gray!10, minimum height=0.5cm, text centered, above=0.2cm of n8.north, anchor=south}] (n9) {Features}; 

\node[style={above=0.2cm of txt3.north, anchor=south, minimum height=0.5cm}] (txt4) {\footnotesize{\textit{(F)}}};


\node[style={rectangle, draw=black, fill=gray!40, minimum height=0.5cm, text centered, anchor=west}] (n10) at (0.5,-4) {Counterfactuals};

\node[style={rectangle, draw=black, fill=gray!40, minimum height=0.5cm, text centered, above=0.2cm of n10.north, anchor=south}] (n11) {Counterfactuals};

\node[style={right=0cm of n11.east}] (plus4) {+};

\node[style={rectangle, draw=black, fill=gray!40, minimum height=0.5cm, text centered, right=0cm of plus4}] (n12) {Feature Importance};

\node[style={right=0cm of n12.east}] (plus5) {+};

\node[style={rectangle, draw=black, fill=gray!40, minimum height=0.5cm, text centered, right=0cm of plus5}] (n13) {Features};

\node[style={rectangle, draw=black, fill=gray!40, minimum height=0.5cm, text centered, above=0.2cm of n12.north, anchor=south}] (n14) {Feature Importance}; 

\node[style={right=0cm of n14.east}] (plus6) {+};

\node[style={rectangle, draw=black, fill=gray!40, minimum height=0.5cm, text centered, right=0cm of plus6}] (n15) {Features}; 

\node[style={rectangle, draw=black, fill=gray!40, minimum height=0.5cm, text centered, above=0.2cm of n15.north, anchor=south}] (n16) {Features}; 



\draw[->, -latex] (n0) -| (n1);
\draw[->, ultra thick, -latex] (n0) -| (n2);

\draw[->, -latex] (n1.west) -- (-8.25,-1) |- (n9.west);
\draw[->, -latex] (n1.west) -- (-8.25,-1) |- (n8.west);
\draw[->, -latex] (n1.west) -- (-8.25,-1) |- (n6.west);
\draw[->, -latex] (n1.west) -- (-8.25,-1) |- (n3.west);

\draw[->, ultra thick, -latex] (n2.east) -- (8.25,-1) |- (n16.east);
\draw[->, ultra thick, -latex] (n2.east) -- (8.25,-1) |- (n15.east);
\draw[->, ultra thick, -latex] (n2.east) -- (8.25,-1) |- (n13.east);
\draw[->, -latex] (n2.east) -- (8.25,-1) |- (n10.east);

\end{tikzpicture}
}
\caption{Graphical representation of our study setup. Thick lines indicate the subset of conditions from our pilot study.
}
\label{fig:study_setup}
\end{figure*}

With our study, we aim to contribute novel insights towards answering the following main questions:

\begin{itemize}
    \item[\textbf{Q1}] Do people perceive a decision process to be fairer and/or more trustworthy if more information about it is disclosed?
    \item[\textbf{Q2}] Does people's experience / knowledge in the field of AI have an impact on their perceptions of fairness and trustworthiness towards automated decision making?
    \item[\textbf{Q3}] How do people perceive human versus automated (consequential) decision making with respect to fairness and trustworthiness? 
\end{itemize}
We choose to explore the aforementioned relationships in the context of lending---an example of a provider-customer encounter.
Specifically, we confront study participants with situations where a person was denied a loan.
We choose a between-subjects design with the following conditions: First, we reveal that the loan decision was made by a human or an ADS (i.e., automated).
Then we provide one of four explanation styles to each study participant.
Figure \ref{fig:study_setup} contains an illustration of our study setup, the elements of which will be explained in more detail shortly.
Eventually, we measure four different constructs: \emph{understandability} (of the given explanations), \emph{procedural fairness} \cite{colquitt2001justice}, \emph{informational fairness} \cite{colquitt2001justice}, and \emph{trustworthiness} (of the decision maker); and we compare the results across conditions.
Additionally, we measure \emph{AI literacy} of the study participants.
Please refer to Appendix \ref{appendix:constructs_items} for a list of all constructs and associated measurement items for the case of automated decisions.
Note that for each construct we measure \emph{multiple} items.

Our analyses are based on a publicly available dataset on home loan application decisions\footnote{\url{https://www.kaggle.com/altruistdelhite04/loan-prediction-problem-dataset} (last accessed Jan 3, 2021)}, which has been used in multiple \texttt{Kaggle} competitions.
Note that comparable data---reflecting a given finance company's individual circumstances and approval criteria---might in practice be used to train ADS.
The dataset at hand consists of 614 labeled (loan Y/N) observations and includes the following features: \textit{applicant income, co-applicant income, credit history, dependents, education, gender, loan amount, loan amount term, marital status, property area, self-employment}.
After removing data points with missing values, we are left with 480 observations, 332 of which (69.2\%) involve the positive label (Y) and 148 (30.8\%) the negative label (N). 
We use 70\% of the dataset for training purposes and the remaining 30\% as a holdout set.

As groundwork, after encoding and scaling the features, we trained a random forest classifier with bootstrapping to predict the held-out labels, which yields an out-of-bag accuracy estimate of 80.1\%.
Our first explanation style, \textit{(F)}, consists of disclosing the features including corresponding values for an observation (i.e., an applicant) from the holdout set whom our model denied the loan.
We refer to such an observation as a \emph{setting}.
%
In our study, we employ different settings in order to ensure generalizability.
Please refer to Appendix \ref{appendix:explanation_styles} for an excerpt of questionnaires for one exemplary setting (male applicant).
Note that all explanations are derived from the data---they are \emph{not} concocted.
Next, we computed permutation feature importances \citep{breiman2001random} from our model and obtained the following hierarchy, using ``$\succ$'' as a shorthand for ``is more important than'': \textit{credit history} $\succ$ \textit{loan amount} $\succ$ \textit{applicant income} $\succ$ \textit{co-applicant income} $\succ$ \textit{property area} $\succ$ \textit{marital status} $\succ$ \textit{dependents} $\succ$ \textit{education} $\succ$ \textit{loan amount term} $\succ$ \textit{self-employment} $\succ$ \textit{gender}.
Revealing this ordered list of feature importances in conjunction with \textit{(F)} makes up our second explanation style \textit{(FFI)}.
To construct our third and fourth explanation styles, we conducted an online survey with 20 quantitative and qualitative researchers to ascertain which of the aforementioned features are actionable---in a sense that people can (hypothetically) act on them in order to increase their chances of being granted a loan.
According to this survey, the top-5 actionable features are: \textit{loan amount, loan amount term, property area, applicant income, co-applicant income}.
Our third explanation style \textit{(FFICF)} is then---in conjunction with \textit{(F)} and \textit{(FFI)}---the provision of three counterfactual scenarios where one actionable feature each is (minimally) altered such that our model predicts a loan approval instead of a rejection.
The last explanation style is \textit{(CF)}, without additionally providing features or feature importances.
This condition aims at testing the effectiveness of counterfactual explanations in isolation, as opposed to providing them in conjunction with other explanation styles.
We employ only model-agnostic explanations \cite{adadi2018peeking} in a way that they could plausibly be provided by both humans and ADS.
%
%
%
%
%
%


\section{Preliminary Analyses and Findings}

%
\begin{table*}
\caption{Pearson correlations between constructs for pilot study.}
\label{tab:cor}
    \begin{tabular}{l l c}
        \toprule
        \textbf{Construct 1} & \textbf{Construct 2} & \textbf{Pearson's $\boldsymbol{r}$}
        \\
        \midrule
        Procedural Fairness & Informational Fairness & 0.47
        \\
        \midrule
        Procedural Fairness & Trustworthiness & 0.78
        \\
        \midrule
        Procedural Fairness & Understandability & 0.23
        \\
        \midrule
        Informational Fairness & Trustworthiness & 0.72 
        \\
        \midrule
        Informational Fairness & Understandability & 0.69
        \\
        \midrule
        Trustworthiness & Understandability & 0.41
        \\
        \bottomrule
    \end{tabular}
\end{table*}
Based on Section \ref{sec:study_design}, we conducted an online pilot study with 58 participants to infer preliminary insights regarding \textbf{Q1} and \textbf{Q2} and to validate our study design.
Among the participants were 69\% males, 29\% females, and one person who did not disclose their gender; 53\% were students, 28\% employed full-time, 10\% employed part-time, 3\% self-employed, and 5\% unemployed.
The average age was 25.1 years, and 31\% of participants have applied for a loan before.
%
%
For this pilot study, we only included the ADS settings (right branch in Figure \ref{fig:study_setup}) and limited the conditions to \textit{(F)}, \textit{(FFI)}, and \textit{(FFICF)}.
%
The study participants were randomly assigned to one of the three conditions, and each participant was provided with two consecutive questionnaires associated with two different settings---one male and one female applicant.
%
%
Participants for this online study were recruited from all over the world via \texttt{Prolific}\footnote{\url{https://www.prolific.co/}} \citep{palan2018prolific} and asked to rate their agreement with multiple statements on 5-point Likert scales, where a score of 1 corresponds to ``strongly disagree'', and a score of 5 denotes ``strongly agree''.
Additionally, we included multiple open-ended questions in the questionnaires to be able to carry out a qualitative analysis as well.
%

\subsection{Quantitative Analysis}
\paragraph{\textbf{Constructs}}
As mentioned earlier, we measured four different constructs: understandability (of the given explanations), procedural fairness \cite{colquitt2001justice}, informational fairness \cite{colquitt2001justice}, and trustworthiness (of the decision maker); see Appendix \ref{appendix:constructs_items} for the associated measurement items.
Note that study participants responded to the same (multiple) measurement items per construct, and these measurements were ultimately averaged to obtain one score per construct.
%
%
%
%
We evaluated the reliability of the constructs through Cronbach's alpha---all values were larger than 0.8 thus showing good reliability for all constructs \citep{cortina1993coefficient}. 
%
%
We proceeded to measure correlations between the four constructs with Pearson's $r$ to obtain an overview of the relationships between our constructs.
Table \ref{tab:cor} provides an overview of these relationships: Procedural fairness and informational fairness are each strongly correlated with trustworthiness, and informational fairness is strongly correlated with understandability.
Overall, we found significant correlations ($p < 0.05$) between all constructs besides procedural fairness and understandability.


\paragraph{\textbf{Insights regarding \normalfont{\textbf{Q1}}}}
We conducted multiple ANOVAs followed by Tukey's tests for post-hoc analysis to examine the effects of our three conditions.
The individual scores for each construct and condition are provided in Table \ref{tab:means}.
We found a significant effect between different conditions on fairness perceptions for procedural fairness ($F(2,55) = 3.56, p = 0.035$) as well as for informational fairness ($F(2,55) = 10.90, p < 0.001$).
Tukey's test for post-hoc analysis showed that the effect for procedural fairness was only significant between the conditions \emph{(F)} and \emph{(FFICF)} ($p = 0.040$).
When controlling for different variables, such as study participants' gender, the effect for procedural fairness is reduced to marginal significance
($p > 0.05$).
%
%
For informational fairness the effect in the post-hoc analysis without control variables is significant between \emph{(F)} and \emph{(FFICF)} ($p < 0.001$) as well as between \emph{(FFI)} and \emph{(FFICF)} ($p = 0.042$), and it is marginally significant between \emph{(F)} and \emph{(FFI)} ($p = 0.072$).
%
%
Controlling for study participants' gender reduces the significance between \emph{(FFI)} and \emph{(FFICF)} to marginal significance ($p = 0.059$); controlling for study participants' age removes the significance between these two conditions altogether.
%
%

Interestingly, significant effects on understandability between conditions ($F(2,55) = 7.52, p = 0.001$) came from \emph{(F)} and \emph{(FFICF)} ($p = 0.001$) as well as \emph{(F)} and \emph{(FFI)} ($p = 0.020$).
%
%
%
Significant effects of the conditions on trustworthiness ($F(2,55) = 4.94, p = 0.011$) could only be observed between \emph{(F)} and \emph{(FFICF)} ($p = 0.007$).
%
%
In general, we urge to exercise utmost caution when interpreting the quantitative results of our pilot study as the sample size is extremely small.
We hope to generate more reliable and extensive insights with our main study and a much larger number of participants.



\begin{table*}
\caption{Construct scores by condition for pilot study. The scores, ranging from 1 (low) to 5 (high), were obtained by averaging across all measurement items for each construct.}
\label{tab:means}
    \begin{tabular}{l c c c}
        \toprule
        \textbf{Construct} &
        \textbf{\emph{(F)}} & \textbf{\emph{(FFI)}} & \textbf{\emph{(FFICF)}}
        \\
        \midrule
        Understandability & 3.17 & 3.87 & 4.12
        \\
        \midrule
        Procedural Fairness & 3.28 & 3.40 & 3.91
        \\
        \midrule
        Informational Fairness & 2.79 & 3.33 & 3.92
        \\
        \midrule
        Trustworthiness & 2.92 & 3.39 & 3.83
        \\
        \bottomrule
    \end{tabular}
\end{table*}

\paragraph{\textbf{Insights regarding \normalfont{\textbf{Q2}}}}
%
We calculated Pearson's $r$ between each of our fairness measures including trustworthiness and the study participants' AI literacy.
All three measures, procedural fairness ($r = 0.35, p = 0.006$), informational fairness ($r = 0.52, p < 0.001$) and trustworthiness ($r = 0.48, p < 0.001$) demonstrate a significant positive correlation with AI literacy.
%
%
Therefore, within the scope of our pilot study, we found that participants with more knowledge and experience in the field of AI tend to perceive the decision making process and the provided explanations of the ADS at hand to be fairer and more trustworthy than participants with less knowledge and experience in this field.
%
%

\subsection{Qualitative Analysis}
%
In the following, we provide a summary of insightful responses to open-ended questions from our questionnaires.

\paragraph{\textbf{Regarding automated decision making}}
Perhaps surprisingly, many participants approved of the ADS as the decision maker.
They perceived the decision to be less biased and argued that all applicants are treated equally, because the ADS makes its choices based on facts, not based on the likeability of a person:
``\textit{I think that an automated system treats every individual fairly because everybody is judged according to the same rules.}''
Some participants directly compared the ADS to human decision makers:
``\textit{I think that [the decision making procedures] are fair because they are objective, since they are automated. Humans usually [can't] make decisions without bias.}''
Other participants responded with a (somewhat expected) disapproval towards the ADS. 
Participants criticized, for instance, that the decisions ``\textit{are missing humanity in them}'' and how an automated decision based ``\textit{only on statistics without human morality and ethics}'' simply cannot be fair. 
One participant went so far as to formulate positive arguments for human bias in decision making procedures:
``\textit{I do not believe that it is fair to assess anything that greatly affects an individual's life or [livelihood] through an automated decision system. I believe some bias and personal opinion is often necessary to uphold ethical and moral standards.}''
Finally, some participants had mixed feelings because they saw the trade-off between a ``\textit{cold approach}'' that lacks empathy and a solution that promotes ``\textit{equality with others}'' because it ``\textit{eliminates personal bias}''.

\paragraph{\textbf{Regarding explanations}} Study participants had strong opinions on the features considered in the loan decision.
Most participants found \emph{gender} to be the most inappropriate feature.
The comments on this feature ranged from ``\textit{I think the gender of the person shouldn't matter}'' to considering gender as a factor being ``\textit{ethically wrong}'' or even ``\textit{borderline illegal}''.
\emph{Education} and \emph{property area} were named by many participants as being inappropriate factors as well: 
``\textit{I think education, gender, property area [\dots] are inappropriate factors and should not be considered in the decision making process.}''
%
%
On average, the order of feature importance was rated as equally appropriate as the features themselves.
Some participants assessed the order of feature importance in general and came to the conclusion that it is appropriate: 
``\textit{The most important is credit history in this decision and least gender so the order is appropriate.}'' 
At the same time, a few participants rated the order of feature importance as inappropriate, for instance because ``\textit{some things are irrelevant yet score higher than loan term.}''
In the first of two settings, the counterfactual for \emph{property area} was received negatively by some:
``\textit{It shouldn't matter where the property is located.}''
Yet, most participants found the counterfactual explanations
in the second setting
to be appropriate:
``\textit{The three scenarios represent plausible changes the individual could perform
[\dots]}''


%
%
%
%
%
%

\section{Outlook}

The potential of automated decision making and its benefits over purely human-made decisions are obvious.
However, several instances are known where such automated decision systems (ADS) are having undesirable effects---especially with respect to fairness and transparency.
With this work, we aim to contribute novel insights to better understand people's perceptions of fairness and trustworthiness towards ADS, based on the provision of varying degrees of information about such systems and their underlying processes.
Moreover, we examine how these perceptions are influenced by people's background and experience in the field of artificial intelligence.
%
%
%
%
%
%
As a first step, we have conducted an online pilot study and obtained preliminary results for a subset of conditions.
Next, we will initiate our main study with a larger sample size and additional analyses.
For instance, we will also explore whether people's perceptions of fairness and trustworthiness change when the decision maker is claimed to be human (as opposed to purely automated).
We hope that our contribution will ultimately help in designing more equitable decision systems as well as stimulate future research on this important topic.

\bibliography{bibliography}

\onecolumn
\appendix

\section{Constructs and Items for Automated Decisions}\label{appendix:constructs_items}

All items within the following constructs were measured on a 5-point Likert scale and mostly drawn (and adapted) from previous studies. 

\begin{enumerate}
    \item \textbf{Understandability}
    
    Please rate your agreement with the following statements:
    \begin{itemize}
        \item The explanations provided by the automated decision system are clear in meaning. \citep{mckinney2002measurement}
        \item The explanations provided by the automated decision system are easy to comprehend. \citep{mckinney2002measurement}
        \item In general, the explanations provided by the automated decision system are understandable for me. \citep{mckinney2002measurement}
    \end{itemize}
    
    \item \textbf{Procedural Fairness}
    
    The statements below refer to the \textit{procedures} the automated decision system uses to make decisions about loan applications. Please rate your agreement with the following statements: 
    
    \begin{itemize}
        \item Those procedures are free of bias. \citep{colquitt2015measuring}
        \item Those procedures uphold ethical and moral standards. \citep{colquitt2015measuring}
        \item Those procedures are fair.
        \item Those procedures ensure that decisions are based on facts, not personal biases and opinions. \citep{colquitt2015measuring}
        \item Overall, the applying individual is treated fairly by the automated decision system. \citep{colquitt2015measuring}
    \end{itemize}
    
    \item \textbf{Informational Fairness}
    
    The statements below refer to the \textit{explanations} the automated decision system offers with respect to the decision-making procedures. Please rate your agreement with the following statements:
    
    \begin{itemize}
        \item The automated decision system explains decision-making procedures thoroughly. \citep{colquitt2015measuring}
        \item The automated decision system’s explanations regarding procedures are reasonable. \citep{colquitt2015measuring}
        \item The automated decision system tailors communications to meet the applying individual’s needs. \citep{colquitt2015measuring}
        \item I understand the process by which the decision was made. \citep{binns2018s}
        \item I received sufficient information to judge whether the decision-making procedures are fair or unfair.
    \end{itemize}
    
    \item \textbf{Trustworthiness}
    
    The statements below refer to the \textit{automated decision system}. Please rate your agreement with the following statements:
    
    \begin{itemize}
        \item Given the provided explanations, I trust that the automated decision system makes good-quality decisions. \citep{lee2018understanding}
        \item Based on my understanding of the decision-making procedures, I know the automated decision system is not opportunistic. \citep{chiu2009understanding}
        \item Based on my understanding of the decision-making procedures, I know the automated decision system is trustworthy. \citep{chiu2009understanding}
        \item I think I can trust the automated decision system. \citep{carter2005utilization}
        \item The automated decision system can be trusted to carry out the loan application decision faithfully. \citep{carter2005utilization}
        \item In my opinion, the automated decision system is trustworthy. \citep{carter2005utilization}
    \end{itemize}
    
    \item \textbf{AI Literacy}
    
    \begin{itemize}
        \item How would you describe your knowledge in the field of artificial intelligence?
        \item Does your current employment include working with artificial intelligence?
    \end{itemize}
    
    Please rate your agreement with the following statements:
    
    \begin{itemize}
        \item I am confident interacting with artificial intelligence. \citep{wilkinson2010construction}
        \item I understand what the term \emph{artificial intelligence} means.
    \end{itemize}
    
\end{enumerate}

\section{Explanation Styles for Automated Decisions and One Exemplary Setting (Male Applicant)}\label{appendix:explanation_styles}

\begin{tcolorbox}[breakable, enhanced jigsaw, sharp corners, opacityback=0, fonttitle=\bfseries, colframe=black, boxrule=1pt, title=Explanation Style \textit{(F)}]

\begin{tcolorbox}[enhanced jigsaw, sharp corners, colframe=black, boxrule=1pt]
A finance company offers loans on real estate in urban, semi-urban and rural areas. A potential customer first applies online for a specific loan, and afterwards the company assesses the customer's eligibility for that loan.

An individual applied online for a loan at this company. The company denied the loan application. The decision to deny the loan was made by an automated decision system and communicated to the applying individual electronically and in a timely fashion.
\end{tcolorbox}

\tcblower

The automated decision system explains that the following factors (in alphabetical order) on the individual were taken into account when making the loan application decision:

\begin{itemize}
    \item Applicant Income: \$3,069 per month
    \item Co-Applicant Income: \$0 per month
    \item Credit History: Good
    \item Dependents: 0
    \item Education: Graduate
    \item Gender: Male
    \item Loan Amount: \$71,000
    \item Loan Amount Term: 480 months
    \item Married: No
    \item Property Area: Urban
    \item Self-Employed: No
\end{itemize}
\end{tcolorbox}

\begin{tcolorbox}[breakable, enhanced jigsaw, sharp corners, opacityback=0, fonttitle=\bfseries, colframe=black, boxrule=1pt, title=Explanation Style \textit{(FFI)}]

\begin{tcolorbox}[enhanced jigsaw, sharp corners, colframe=black, boxrule=1pt]
A finance company offers loans on real estate in urban, semi-urban and rural areas. A potential customer first applies online for a specific loan, and afterwards the company assesses the customer's eligibility for that loan.

An individual applied online for a loan at this company. The company denied the loan application. The decision to deny the loan was made by an automated decision system and communicated to the applying individual electronically and in a timely fashion.
\end{tcolorbox}

\tcblower

The automated decision system explains \dots 
\begin{itemize}
    \item \dots that the following factors (in alphabetical order) on the individual were taken into account when making the loan application decision:

    \begin{itemize}
        \item Applicant Income: \$3,069 per month
        \item Co-Applicant Income: \$0 per month
        \item Credit History: Good
        \item Dependents: 0
        \item Education: Graduate
        \item Gender: Male
        \item Loan Amount: \$71,000
        \item Loan Amount Term: 480 months
        \item Married: No
        \item Property Area: Urban
        \item Self-Employed: No
    \end{itemize}
    \item \dots that different factors are of different importance in the decision. The following list shows the order of factor importance, from most important to least important:
    Credit History $\succ$ Loan Amount $\succ$ Applicant Income $\succ$ Co-Applicant Income $\succ$ Property Area $\succ$ Married $\succ$ Dependents $\succ$ Education $\succ$ Loan Amount Term $\succ$ Self-Employed $\succ$ Gender
\end{itemize}
\end{tcolorbox}

\begin{tcolorbox}[breakable, enhanced jigsaw, sharp corners, opacityback=0, fonttitle=\bfseries, colframe=black, boxrule=1pt, title=Explanation Style \textit{(FFICF)}]

\begin{tcolorbox}[enhanced jigsaw, sharp corners, colframe=black, boxrule=1pt]
A finance company offers loans on real estate in urban, semi-urban and rural areas. A potential customer first applies online for a specific loan, and afterwards the company assesses the customer's eligibility for that loan.

An individual applied online for a loan at this company. The company denied the loan application. The decision to deny the loan was made by an automated decision system and communicated to the applying individual electronically and in a timely fashion.
\end{tcolorbox}

\tcblower

The automated decision system explains \dots 
\begin{itemize}
    \item \dots that the following factors (in alphabetical order) on the individual were taken into account when making the loan application decision:

    \begin{itemize}
        \item Applicant Income: \$3,069 per month
        \item Co-Applicant Income: \$0 per month
        \item Credit History: Good
        \item Dependents: 0
        \item Education: Graduate
        \item Gender: Male
        \item Loan Amount: \$71,000
        \item Loan Amount Term: 480 months
        \item Married: No
        \item Property Area: Urban
        \item Self-Employed: No
    \end{itemize}
    \item \dots that different factors are of different importance in the decision. The following list shows the order of factor importance, from most important to least important:
    Credit History $\succ$ Loan Amount $\succ$ Applicant Income $\succ$ Co-Applicant Income $\succ$ Property Area $\succ$ Married $\succ$ Dependents $\succ$ Education $\succ$ Loan Amount Term $\succ$ Self-Employed $\succ$ Gender
    \item \dots that the individual would have been granted the loan if---everything else unchanged---one of the following hypothetical scenarios had been true:
    \begin{itemize}
        \item The Co-Applicant Income had been at least \$800 per month
        \item The Loan Amount Term had been 408 months or less
        \item The Property Area had been Rural
    \end{itemize}
\end{itemize}
\end{tcolorbox}

\end{document}